\documentclass[journal]{IEEEtran}
\usepackage{multirow}
\usepackage{times}
\usepackage{epsfig}
\usepackage{graphicx}
\usepackage{amsmath}
\usepackage{amssymb}
\usepackage{arydshln}
\usepackage{algorithmic}
\usepackage{algorithm}
\usepackage[T1]{fontenc}
\usepackage[utf8]{inputenc}
\usepackage{color}
\usepackage{verbatim}
\begin{document}

\title{Quality-aware Feature Aggregation Network \\for Robust RGBT Tracking }

\author{Yabin Zhu,~~Chenglong Li,~~Jin Tang,~~Bin Luo
\thanks{The authors are with School of Computer Science and Technology, Anhui University, Hefei 230601, China.}}

\markboth{ October~2019}%
{Shell \MakeLowercase{\textit{et al.}}: Bare Demo of IEEEtran.cls for IEEE Journals}

\maketitle

\begin{abstract}
This paper investigates how to perform robust visual tracking in adverse and challenging conditions using complementary visual and thermal infrared data (RGBT tracking).
We propose a novel deep network architecture called quality-aware Feature Aggregation Network (FANet) for robust RGBT tracking. 
Unlike existing RGBT trackers, our FANet aggregates hierarchical deep features within each modality to handle the challenge of significant appearance changes caused by deformation, low illumination, background clutter and occlusion. 
In particular, we employ the operations of max pooling to transform these hierarchical and multi-resolution features into  uniform space with the same resolution, and use  1$\times$1 convolution operation to compress feature dimensions to achieve more effective hierarchical feature aggregation. 
To model the interactions between RGB and thermal modalities, we elaborately design an adaptive aggregation subnetwork to integrate features from different modalities based on their reliabilities and thus are able to alleviate noise effects introduced by low-quality sources.
The whole FANet is trained in an end-to-end manner.
Extensive experiments on large-scale benchmark datasets demonstrate the high-accurate performance against other state-of-the-art RGBT tracking methods. 
\end{abstract}


%
\IEEEpeerreviewmaketitle

\section{Introduction}
\begin{figure*}[t]
  \centering
  \includegraphics[width=\textwidth]{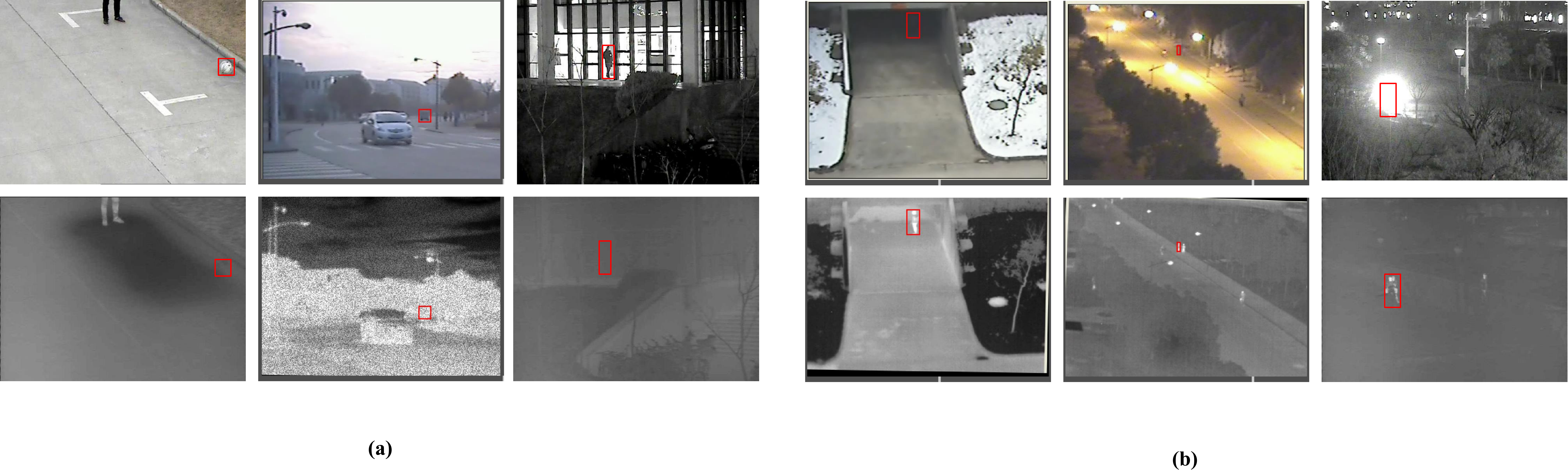} \\
  \caption{ Typical complementary cases of RGB and thermal data. (a) RGB data source is better than thermal ones, where
thermal spectrum is disturbed by thermal crossover, noise and glass. (b)
Thermal sources better than RGB ones, where visible spectrum is disturbed by low illumination and high illumination.  }\label{fig::challege}
\end{figure*}

 Thermal cameras recently become more economically affordable and have been applied to many computer vision tasks, such as object segmentation~\cite{Li17weld, li2019segmenting}, person Re-ID~\cite{Wu17iccv} and pedestrian detection~\cite{Hwang15cvpr,Xu17cvpr}, to name a few. 
They are insensitive to lighting condition and have a strong ability to penetrate  smog and haze~\cite{Gade14mva}, and thus able to provide strong complementary information for visible cameras. 
However, thermal infrared data are limited by the challenge of thermal crossover and lacking in color and texture details.
Visible spectrum information can make up for it, and Fig.~\ref{fig::challege} shows some examples. 
Therefore, effective fusion of visible and thermal data has big potential in handling various challenges for robust visual tracking. 

As a subbranch of visual tracking, RGBT tracking aims to take advantage of complementary information from visible and thermal infrared spectra to estimate the state of a specific instance in a video, given the ground truth bounding box in the first frame. 
Recent studies on RGBT tracking mainly focus on two aspects. 
The one is aiming to learn robust feature representation via the usage of RGB and thermal data~\cite{Li17rgbt210,li18nuecom,Li18eccv} and achieves promising tracking performance. 
These works rely on either handcrafted features or highly semantic deep features. 
Handcrafted features are too weak to handle the challenges of significant appearance changes caused by deformation, low illumination, background clutter and occlusion within each modality. 
Although highly semantic features are effective to distinguish target semantics,
they are unable to meet the purpose of visual tracking, which is to locate objects accurately rather than just to infer their semantic categories. 
Therefore, relying on handcrafted features or highly semantic features only is difficult to handle various challenges in RGBT tracking.

The another one focuses on introducing modality weights to achieve adaptive aggregation of different source data~\cite{Li16tip,lan18aaai}. 
Early works employ the reconstruction residues~\cite{Li16tip,Li17rgbt210} to regularize modality weight learning. 
Lan \emph{et al.}~\cite{lan18aaai} uses the max-margin principle to optimize the modality weights according to classification scores. 
However, these works would fail when reconstruction residuals or classification scores do not reliably reflect modal reliability. 
For example, for the $BlurCar$ sequence in GTOT dataset~\cite{Li16tip}, the quality of RGB modality is much better than thermal modality. 
But the reconstruction residue on RGB modality computed by the model in~\cite{Li16tip} is 37.60, which is higher than the residue of thermal modality 32.67. 
Therefore, the uncertainty of reliability degree in different modalities is still an uncrossed obstacle for these type of works. 

To solve these problems, we propose a novel architecture, namely quality-aware Feature Aggregation Network (FANet), for robust RGBT tracking. 
Within each modality, our FANet first aggregates hierarchical multi-resolution feature maps into a uniform space at the same resolution. 
Shallow features encode fine-grained spatial details such as target positions and help achieving precise target localizations, while deep features are more effective to capture target semantics. 
Existing methods~\cite{Ma15iccv,HDT16cvpr,C-COT16eccv,ECO17cvpr} usually select part of deep features extracted from pre-trained networks to learn correlation filter models for visual tracking. 
However, using only part of features and unchanged network parameters in training and tracking processes would limit tracking performance.
Therefore, we propose an effective method to deploy all layer features and learn robust target representations in an end-to-end trained network.
In a specific, we employ the operations of max pooling to convert their resolutions into uniform feature space, and use the convolution operations to suppress redundant features.

In different modalities, we elaborately design an adaptive aggregation subnetwork to integrate features from different modalities based on their reliability. 
Many efforts are devoted to calculate the modality weights that reflect the reliability degrees to make adaptive aggregation of different modalities, and thus achieve significant improvement for tracking performance~\cite{Li16tip,lan18aaai}. 
In this paper, we come up with a more effective method to compute the modality weights in an end-to-end learning framework. 
Given the aggregated features, we first fuse all modalities feature using a concatenation operation and compress the global spatial information into a channel descriptor utilizing the global average pooling. 
Then, we use a fully connected layer to learn a nonlinear interaction between channels, and the fully connected layers and the SoftMax activation function are used to predict the modality weights that control the information flows of different modalities for adaptive aggregation. 
The object tracking is performed via the binary classification in the multi-domain learning network~\cite{MDNet15cvpr}. 
We summarize our major contributions of our work as follows.

\begin{itemize}
\item We propose a novel end-to-end trained deep network to learn a powerful feature representations for robust RGBT tracking. 
The network aggregates hierarchical deep features effectively and multi-modal information adaptively to handle the challenge of significant appearance changes caused by deformation, low illumination, background clutter and occlusion in RGBT tracking.

\item We design a feature aggregation module and an adaptive aggregation module to fuse multiple features within each modality and in different modalities respectively. 
In particular, the operations of max pooling are employed to transform hierarchical deep features into a uniform space with the same resolution, and the operations of 1$\times$1 convolution are used to compress feature dimensions for more effective hierarchical feature aggregation. In addition, the operations of global average pooling, fully connected layer and SoftMax activation function are used to predict the modality weights for adaptive aggregation.

\item Extensive experiments on large-scale benchmark datasets prove that the FANet significantly better than other state-of-the-art tracking methods with a clear margin and achieves 19 FPS.

\end{itemize}

\begin{figure}[t]
  \centering
  \includegraphics[width=\columnwidth]{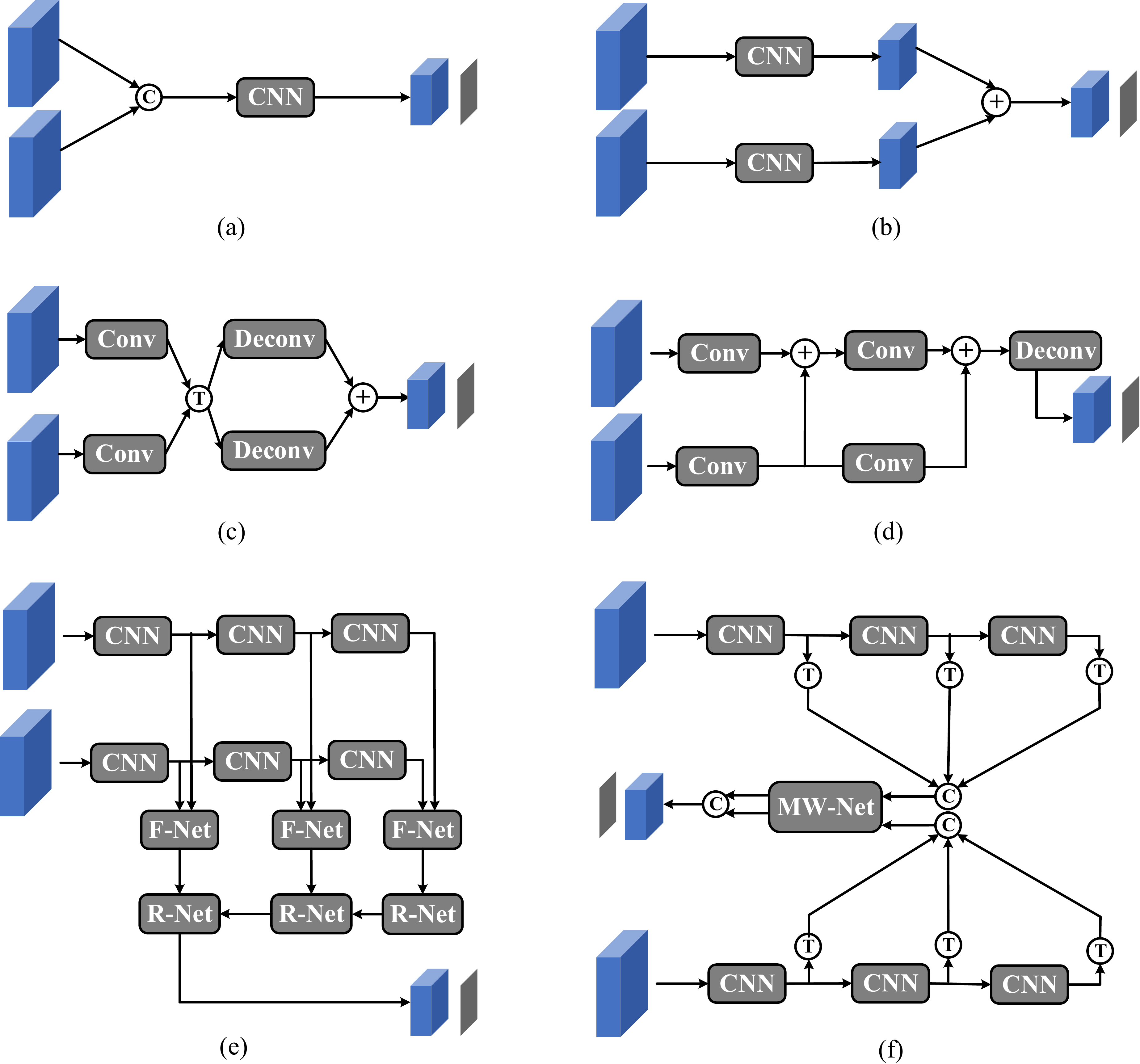} \\
  \caption{Different existing architectures of multimodal fusion. (a) Early fusion. (b) Late fusion. (c) Architecture proposed
by~\cite{network-I16eccv}. (d) Architecture proposed in~\cite{network-II16accv}. (e) Architecture proposed in~\cite{network-III17iccv}. (f) Our architecture. `C', `T' and `+' represent concatenation, transformation, and element-wise summation. And `F-Net', 'R-Net'and `MW-Net' indicate fusion, refinement and modality weight prediction networks, respectively. }\label{fig::Structures}
\end{figure}

\section{Related Work}

\label{sec::related_work}
Based on the relevance to our work, we review the related works from four research lines: RGB-Thermal fusion, hierarchical features for tracking, multi-domain network for tracking, and multimodal aggregation network architectures.

\subsection{RGB-Thermal Fusion for Tracking}
RGBT tracking become more and more attentions in computer vision community with the popularity of thermal infrared sensors~\cite{Li17weld,Wu17iccv,Hwang15cvpr,Xu17cvpr,Li18eccv}.
One research stream is to introduce modality weights to achieve adaptive aggregation of different modalities~\cite{Li17tsmcs,Li16tip,lan18aaai}. 
Early works employ reconstruction residues~\cite{Li17tsmcs,Li16tip} to regularize the modality weight learning and carry out object tracking in Bayesian filtering framework. 
Lan \emph{et al.}~\cite{lan18aaai} use the max-margin principle to optimize the modality weights according to classification scores. 
However, these methods would fail when reconstruction residuals or classification scores do not reliably reflect modal reliability.

The other research stream is to learn robust feature representations via the usage of multimodal data~\cite{Li17rgbt210,li18nuecom,Li18eccv}. 
Li \emph{et al.}~\cite{Li17rgbt210} propose a weighted sparse representation regularized graph learning approach to construct a graph-based multimodal descriptor, and adopt structured SVM for tracking. 
And Li \emph{et al.}~\cite{Li18eccv} further consider the heterogeneous property between different modalities and noise effects of initial seeds in the cross-modal ranking model. 
 Li \emph{et al.}~\cite{li18nuecom}  design a FusionNet that select most discriminative feature maps from deep feature of CNN to avoid feature redundant and noises for adaptive aggregation of different modalities.
These methods rely on hand-crafted features or highly semantically deep features to locate objects, therefore it is difficult to cope with the challenges of significant appearance changes caused by background clutter, low illumination, deformation and occlusion within each modality.

\subsection{Hierarchical Features for Tracking}
In recent literatures, several works started incorporating hierarchical deep features for the task of visual tracking~\cite{Ma15iccv,HDT16cvpr,HyperNet16cvpr,MCPF17cvpr,C-COT16eccv,ECO17cvpr}. 
Ma \emph{et al.}~\cite{Ma15iccv} interprets the hierarchical features of the convolutional layer as a non-linear correspondence of the representation of the image pyramid and explicitly uses these multiple levels of abstraction to represent the target object.
 In order to make full use of the multi-layer CNN features, Qi \emph{et al.}~\cite{HDT16cvpr} designs a Hedge method to hedge multiple trackers into more powerful ones.
To take the interdependencies among different features into account, Zhang \emph{et al.}~\cite{MCPF17cvpr} present a multi-task correlation filter to learn the correlation filters jointly. 
A combination of handcrafted low-level and hierarchical deep features is proposed by Danelljan \emph{et al.}~\cite{C-COT16eccv,ECO17cvpr} by employing an implicit interpolation model to construct the learning problem in the continuous spatial space, which makes multi-resolution feature maps integrated effectively. 
These methods, however, are based on the assumption that multiple features contribute equally or set the feature weights manually, therefore they cannot fully and effectively use features according to their reliabilities.

\begin{figure*}[t]
  \centering
  \includegraphics[width=\textwidth]{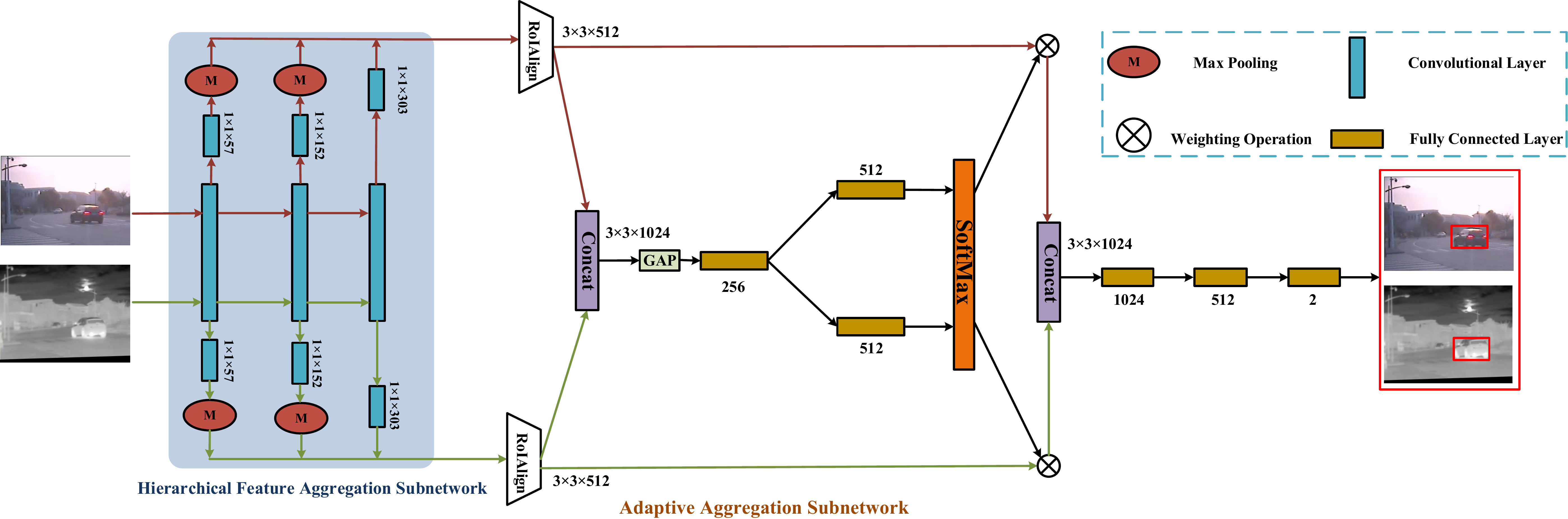} \\
  \caption{Diagram of our FANet architecture that consists of a hierarchical feature aggregation subnetwork and an adaptive aggregation subnetwork. }\label{fig::FANet}
\end{figure*}

\subsection{Multi-Domain Networks for Tracking}
MDNet~\cite{MDNet15cvpr} has achieved the state-of-the-art performance on multiple datasets through multi-domain learning. 
Han \emph{et al.}~\cite{BranchOut17cvpr} proposes a random selection of a branches of CNN for online learning when target object appearance models need to be updated for better regularization, where each of the branches may have a different number of layers to maintain a variable level of abstraction of the target's appearance. 
The meta learning is introduced in MDNet to adjust the initialization of the deep networks~\cite{MetaTracker18eccv}, which could quickly adapt to robustly model a particular target in future frames. 
Jung \emph{et al.}~\cite{RT-MDNet18eccv} propose a real-time MDNet, where an improved RoIAlign technique extracts a more accurate feature representation of the target. 
All these methods are developed for single-modality tracking and we study them in the task of multi-modality tracking in this work.

\subsection{Multimodal Fusion Network Architectures}
Early fusion works simply concatenate input multimodal channels and then adopt the convolutional networks (CNN) to extract feature representations (\emph{i.e.} Fig.~\ref{fig::Structures} (a)). 
Some later fusion works additionally report the result of fusing multiple feature representations extracted from the CNN of each modality~\cite{li18nuecom} (\emph{i.e.} Fig.~\ref{fig::Structures} (b)). 
Wang \emph{et al.}~\cite{network-I16eccv} propose a structure for deconvolution of multiple modalities, in which an additional feature transformation network is introduced to associate two modalities by discovering common and modal-specific features (\emph{i.e.} Fig.~\ref{fig::Structures} (c)). 
It doesn't exploit any abundant intermediate feature of both modalities, and the training procedures are not end-to-end. 
Hazirbas \emph{et al.}~\cite{network-II16accv} proposes to use intermediate modal features because they simply add intermediate multimodal features, so the effective mid-level multimodal features cannot be fully utilized. 
Park \emph{et al.}~\cite{network-III17iccv}proposes a network that can effectively utilize multi-level features simultaneously (\emph{i.e.} Fig.~\ref{fig::Structures} (e)). 
However, it neglect the reliability degrees of different modalities, and thus might be affected by noises of features and modalities. 
In this work, we take all above issues into account and propose a quality-aware feature aggregation network for effective multimodal fusion (\emph{i.e.} Fig.~\ref{fig::Structures} (f)).

\section{FANet Framework}
In this section, we introduce the details of our FANet (quality-aware Feature Aggregation Network) framework, including network architecture, training procedure and tracker details.

\subsection{Network Architecture}
The overall network architecture of our FANet is shown in Fig.~\ref{fig::FANet}. 
FANet mainly consists of two parts, i.e., a hierarchical feature aggregation subnetwork and an adaptive aggregation subnetwork. 
Followed by these two subnetworks, we add a binary classification layer with three fully connected layers and a SoftMax with Binary Cross Entropy (BCE) loss to carry out tracking task. 
The three fully connected layers (with 1024, 512, 2 output units respectively (see Fig.~\ref{fig::FANet})) are combined
with ReLUs and dropouts. 
The third fully connected layer with two output units include $K$ branches, where each branch corresponds to a specific domain that is a video sequence for the tracking task. 
The last layer is a SoftMax with BCE loss to perform binary classification to distinguish the foreground(target object) and the background in each domain. 
And in the training phase, the instance embedding loss~\cite{RT-MDNet18eccv} function is also adopted to learn more discriminative target representation models.
%
%
Unlike the feature extraction of MDNet~\cite{MDNet15cvpr}, we extract the features of samples directly from the feature maps through a RoIAlign operation~\cite{RT-MDNet18eccv}, which greatly reduces the computational burden and accelerates the running speed of FANet.

\begin{figure}[t]
  \centering
  \includegraphics[width=\columnwidth]{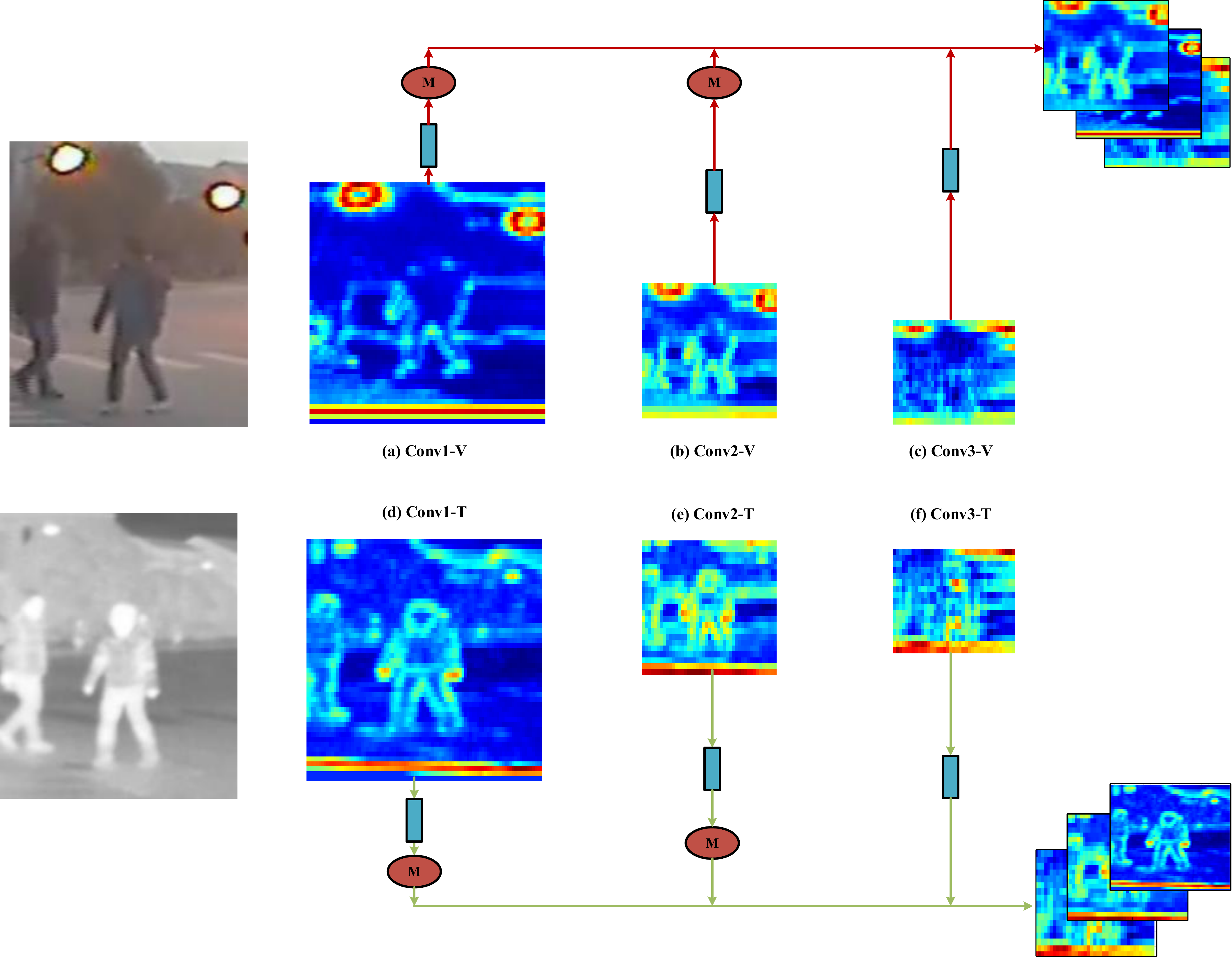} \\
  \caption{Illustration of hierarchical feature aggregation. }\label{fig::features-aggregation}
\end{figure}

\subsection{Hierarchical Feature Aggregation Subnetwork}
Feature aggregation plays an important role in improving tracking performance since diverse features are able to provide complementary advantages which could handle various challenges.
However, existing methods~\cite{Ma15iccv,HDT16cvpr,C-COT16eccv,ECO17cvpr} usually select part of deep features extracted from pre-trained networks to learn correlation filter models for visual tracking. 
There are two major limitations. 
1) They are unable to make the best use of hierarchical features since only part of features are employed.
2) These features are hard to handle all challenges well because the parameters of the used networks remain unchanged in training and tracking processes.
To handle these problems, we propose an effective approach to aggregate the shallow-to-deep features of all layers in an end-to-end trained network, as shown in Fig.~\ref{fig::FANet} and Fig.~\ref{fig::features-aggregation}. 

In a specific, we extract hierarchical features (\emph{Conv1}-\emph{Conv3}) using the improved VGG-M Network~\cite{vgg15iclr} pre-trained on the ImageNet dataset for informative target representations. In order to increase the receptive field and improve quality of representations of RoI,  we refine the VGG network. Specifically, we remove the max pooling layer followed by \emph{conv2} layer and modify \emph{conv3} to be a dilated convolutions~\cite{chen2017deeplab} with rate $r=3$.
Features in different layers have different resolutions due to pooling operations. 
Therefore, to convert these features to the same resolution, we employ the max pooling operation to subsample high-resolution features. 
Furthermore, to transform hierarchical features into a uniform space and compress features, we add a convolution operation for each layer and adopt a local response normalization (LRN) layer to normalize these feature maps.
It is worth noting that different modalities share the parameters in the first three convolutional layers (i.e., \emph{Conv1}-\emph{Conv3}). 

That's because different modalities should share a large number of parameters to represent their collaboration information, and the shared structure can avoid many redundant parameters.
Features of different layers should make different contributions to some certain video sequences as shallow features preserve the fine-grained details while deep features capture target semantics. 
Therefore, our feature aggregation subnetwork aggregates the spatial and semantic information of the two modalities from shallow to deep, and also compresses the feature channels so that more rich and effective feature representation can be obtained. 
Fig.~\ref{fig::FANet} shows the details of the subnetwork configuration and partial parameters. 



\subsection{Adaptive Aggregation Subnetwork}
Duo to different imaging properties of RGB and thermal sensors, their reliability degrees, representing the contributions to the tracking performance, should be different. 
Several works attempt to introduce modality weights to achieve adaptive aggregation of different source data~\cite{Li16tip,Li17rgbt210,lan18aaai}. 
They employ the reconstruction residues~\cite{Li16tip,Li17rgbt210} or classification scores~\cite{lan18aaai} to regularize modality weight learning, and thus would fail when reconstruction residuals or classification scores do not reliably reflect modal reliability.

In this work, inspired by ~\cite{li2019selective}, we elaborately design an adaptive aggregation subnetwork to predict the modality weights in an end-to-end trained network. 
Note that our motivation is significantly different from ~\cite{li2019selective} from the following two aspects. 
First,~\cite{li2019selective} presents a nonlinear approach to aggregate information from multiple kernels to realize the adaptive receptive field sizes of neurons. 
However, we are trying to adaptively aggregate the information of different modalities, and do not consider the different receptive fields of the same feature layer.
Second,~\cite{li2019selective} aggregates the feature maps of different receptive fields with additive operation, while we integrate different modalities with concatenate operation, because the addition operation is likely to cause negative effects when the quality of some modal features is low. 
Our adaptive aggregation subnetwork has two steps including RGBT feature embedding and adaptive weighting.

\begin{figure}[t]
  \centering
  \includegraphics[width=\columnwidth]{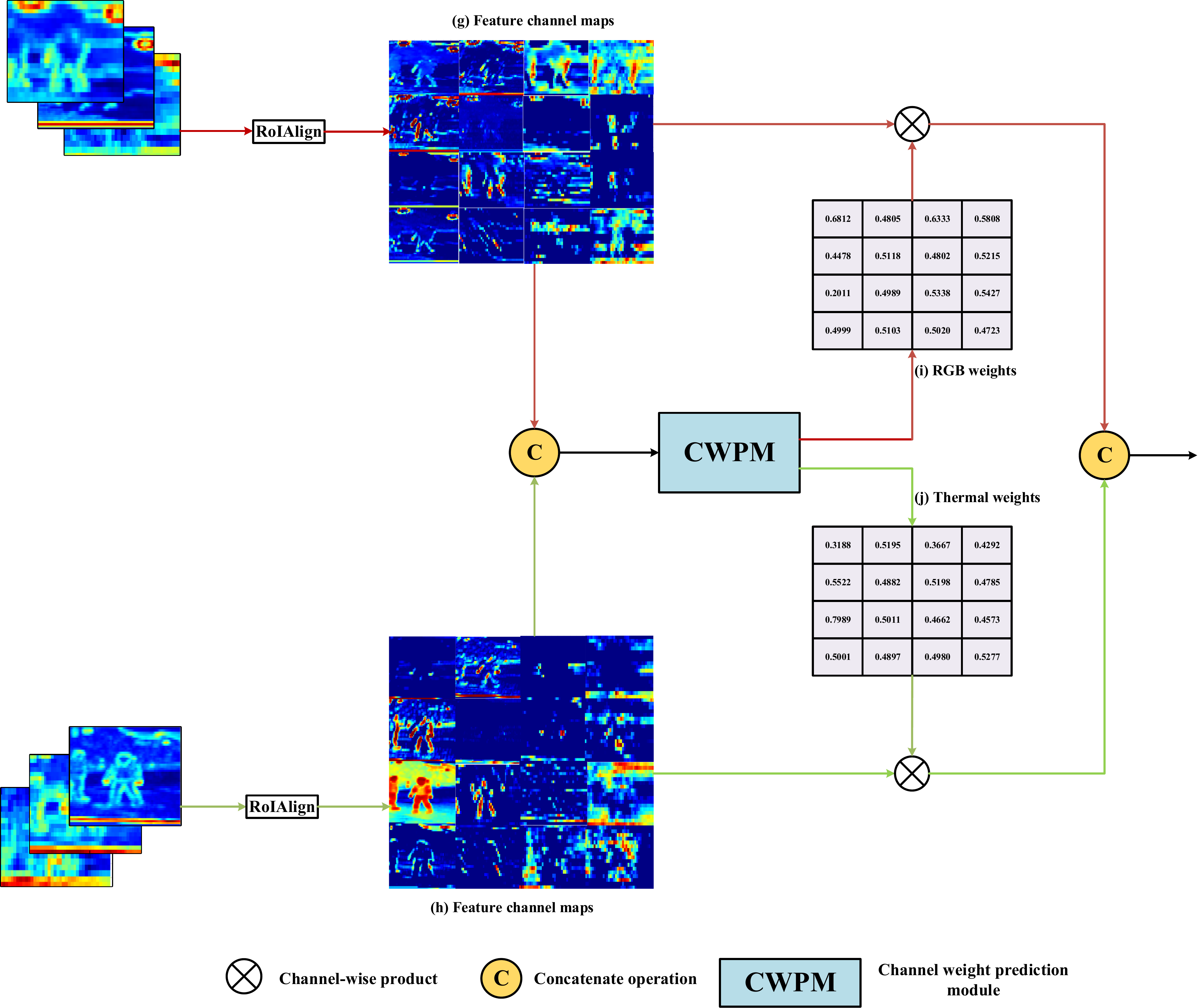} \\
 \caption{Illustration of adaptive aggregation. (g) and (h) depict the feature channel maps, (i) and (j) denote the channel weights calculated from  the feature maps. 
The details of the channel weight prediction module (CWPM) can be found in Fig.~\ref{fig::FANet}
 }\label{fig::modality-aggregation}
\end{figure}

{\flushleft \bf RGBT feature embedding}. 
We aim to adaptively integrate features from RGB and thermal sources according to their reliabilities to tracking performance. 
The basic thought is to utilize the gate mechanism to control the information flows form different modalities. 
To this end, we first fuse all modality features (shown in  Fig.~\ref{fig::FANet} and Fig.~\ref{fig::modality-aggregation}) using a concatenate operation. 
Then we compress global spatial information into a channel descriptor utilizing the global average pooling. Specifically, the $c$-th element of the compressed features $f$ is calculated by:
\begin{equation}
\label{eq::GAP}
f_{c} =\frac{1}{{ W}\times{ H}} \sum_{j=1}^{W}\sum_{k=1}^{H} {X}_c(j,k)
\end{equation}
where $X=\left[X_{1}, \cdots, X_{C}\right] \in \mathbb{R}^{C \times H \times W}$ is the input feature maps, and  $X=\left[X_{rgb}, X_{t}\right]$.
$X_{rgb}$ and $X_{t}$ are the aggregated feature maps of RGB and thermal sources, respectively. 
$C$ denotes the number of channels, and $H$ and $W$ denote the height and width of feature maps respectively. 

{\flushleft \bf Adaptive weighting}.
A compact features $\mathbf{z} \in \mathbb{R}^{d \times 1} $ are created to learn nonlinear interactions between channels. 
This is done through a fully connected layer with the reduction of dimensionality for better efficiency:
\begin{equation}
\mathbf{z}=\mathcal{F}(f)=\delta(Wf)
\end{equation}
where $\delta$ denotes the $ReLU$ function, and $\mathcal{F}(\cdot)$ indicates a fully connected operation. 
$W \in \mathbb{R}^{d \times C}$ denotes the filter of the fully connected layer, and in this paper $d$ is set to 256.

Further, to adaptively aggregate information of different modalities, the fully connected layer and the SoftMax activation function are used. 
Note that we use a fully connected layer for each modality to weigh different modality features as follows:
\begin{equation}
\begin{split}
\mathbf{V}_{rgb}=\delta(\mathbf{W}_{21} \mathbf{z}),\mathbf{V_{t}}=\delta(\mathbf{W}_{22} \mathbf{z}),\\
a_{c}=\frac{e^{\mathbf{V}_{rgb}^{c}}}{e^{\mathbf{V}_{rgb}^{c}}+e^{\mathbf{V}_{t}^{c}}}, b_{c}=\frac{e^{\mathbf{V}_{t}^{c}}}{e^{\mathbf{V}_{rgb}^{c}}+e^{\mathbf{V}_{t}^{c}}}
\end{split}
\end{equation}
where $\mathbf{V}_{rgb},\mathbf{V}_{t} \in \mathbb{R}^{C \times 1}$ and $ a,b $ denote the attention vector for $X_{rgb}$ and $ X_{t}$, respectively. 
$\mathbf{V}_{rgb}^{c},\mathbf{V}_{t}^{c} \in \mathbb{R}^{1 \times 1}$ and $a^{c},b^{c} $ are the $c-\mathrm{th}$ row of $\mathbf{V}_{rgb},\mathbf{V}_{t}$ and $a, b$. $\mathbf{W}_{21},\mathbf{W}_{22} \in \mathbb{R}^{C \times d}$ is filter for the fully connected layer operation.

Finally, the aggregated feature maps can be obtained:
\begin{equation}
\mathbf{F}_{c}=a_{c} \cdot {X}_{rgb}^{c}+b_{c} \cdot {X}_{t}^{c}
\end{equation}
where $\mathbf{F}=\left[\mathbf{F}_{1}, \mathbf{F}_{2}, \ldots, \mathbf{F}_{C}\right], \mathbf{F}_{c} \in \mathbb{R}^{H \times W}$.

\subsection{Training Procedure}
The whole network is trained in an end-to-end manner. 
We first initialize the parameters of the first three convolutional layers using the pre-trained model of the VGG-M network~\cite{vgg15iclr}. 
Then, we train the whole network by the Adam optimization algorithm where each domain is handled separately. 
The detailed settings of training are presented as follows. 
In each iteration, a minibatch is constructed from 8 frames which are randomly chosen in each video sequence. 
And we draw 32 positive and 96 negative samples from each frame and generate 256 positive and 768 negative samples together in a minibatch. 
Herein, the samples whose the IoU overlap ratios with the ground-truth bounding box are larger than 0.7 are treated as positive, and the negative samples have less than 0.5 IoUs. 
For multi-domain learning with $K$ training sequences, we train the network with $200\times K$ iterations, where the learning rate is 0.0001 for convolutional layers and 0.001 for fully connected layers. 
The weight decay is fixed to 0.0005, respectively. 
We train our network on all video sequences from RGBT234 dataset~\cite{li2019rgb} and test it on GTOT dataset~\cite{Li16tip}. 
For another experiment, we train our network on all video sequences from GTOT dataset and test it on RGBT234 dataset.

In this training phase, the loss function of binary cross entropy (BCE) is used for binary classification (foreground and background) for each domain (\emph{i.e.} each video sequence), and the BCE loss can be formulated as:
\begin{equation}
\mathcal{L}_{cls}=-\frac{1}{N}\sum_{i=1}^{N} y_{i} \log p_{i}+\left(1-y_{i}\right) \log \left(1-p_{i}\right)
\end{equation}
where $N$ is the number of samples, and $p_{i}$ is the predicted probabilistic value of the $i$-th sample from FANet. 
$y_{i}$ is the ground truth label of corresponding sample, where $y_{i}$ is 1 for positive samples, otherwise 0.

In addition, the loss function of instance embedding~\cite{RT-MDNet18eccv} is also adopted to learn a more discriminative embedding of target objects with similar semantics. 
Unlike BCE loss, this instance embedding loss is for all domains and treats each domain as a category, and only positive samples of each domain are used in calculating this loss. 
The instance embedding loss can learn discriminative representations of the unseen target objects in new test sequences by forcing target objects in different sequences to be embedded far away from each other.  
It can be formulated as :
\begin{equation}
\mathcal{L}_{inst} = -\frac{1}{N} \sum_{i=1}^{N} \sum_{d=1}^{D} y_{i, d} \log p_{i,d}
\end{equation}
where $D$ is the number of domains, and $y_{i,d}$ is the ground truth label of the $i$-th sample of the $d$-th domain. 
$p_{i,d}$ is the predicted probabilistic value of the $i$-th sample of the  $d$-th domain from FANet.

The final loss function for the optimization of our FANet is formulated as:
\begin{equation}
\mathcal{L}=\mathcal{L}_{\mathrm{cls}}+\alpha \cdot \mathcal{L}_{\mathrm{inst}}
\end{equation}
where $\alpha$ is a hyper-parameter that controls the balance between the two loss terms and in this paper we set it to $\alpha=0.1$.

\subsection{Tracker Details}
In tracking, we replace the $K$ branches of domain specific (the last fc layer) with a single branch for each test sequence. 
The newly added branch is trained in the first frame pair and updated in subsequent frame pairs. 
Given the first frame pair with the ground truth of target object, we draw 500 positive (IoU with ground truth is larger than 0.7) and 5000 negative samples (IoU with ground truth is smaller than 0.5), and train the new branch with 30 iterations, where the learning rate of the last fc layer is set to 0.001 and others are 0.0001. 
Given the $t$-th frame, we draw a set of candidates $\{{\bf x}_t^i\}$ from a Gaussian distribution of the previous tracking result ${\bf x}^*_{t-1}$, where the mean of Gaussian function is set to ${\bf x}^*_{t-1}=(x_{t-1},y_{t-1},s_{t-1})$ and the covariance is set as a diagonal matrix $diag\{0.09r^2,0.09r^2,0.25\}$. 
$(x,y)$ and $s$ indicate the location and scale respectively and $r$ is the mean of $(x_{t-1},y_{t-1})$. 
For the $i$-th candidate ${\bf x}_t^i$, we compute its positive and negative scores using the trained network as $f^+({\bf x}_t^i)$ and $f^-({\bf x}_t^i)$, respectively. 
The target location of current frame is the candidate with the maximum positive score as:
\begin{equation}
\label{edge:score}
{\bf x}_t^* = \arg\max_{{\bf x}_t^i} f^+({\bf x}_t^i),~i=1,2,...,N,
\end{equation}
where $N$ is the number of candidates. 
We also apply bounding box regression technique~\cite{MDNet15cvpr} to improve target localization accuracy. 
We only train the regenerator in the first frame to avoid potential unreliability of other frames. 
If the estimated target state is reliable enough, \emph{i.e.}, $f^+({\bf x}_t^*)>0.5$, we adjust the target locations using the regression model. 
During the tracking process, our network has a long-term update strategy and a short-term update strategy. 
When $f^+({\bf x}_t^i)<0$, the short-term update is used, and the last 20 positive samples and 20 negative samples are adapted to fine-tune network parameters. 
Correspondingly, a long-term update is performed at 10 frames interval, and the last 100 positive samples and 20 negative samples are adapted to fine-tune the parameters of our tracker. 
To reduce the computational burden and avoid over-fitting, we only update the network parameters of fully connected layers.

\begin{figure*}[t]
  \centering
  \includegraphics[width=\textwidth]{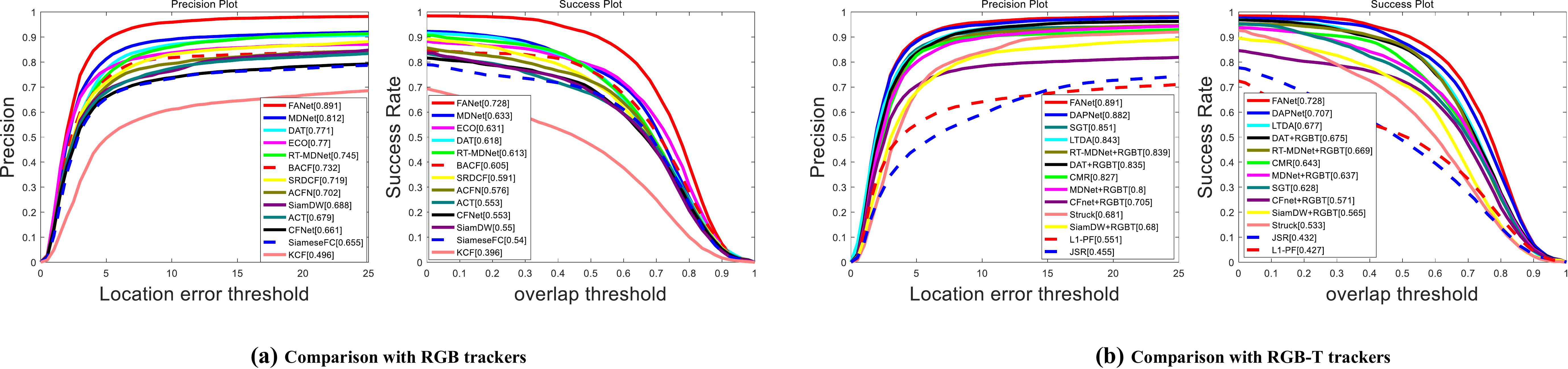} \\
  \caption{The evaluation results on GTOT dataset. The representative scores of PR/SR is presented in the legend. }\label{fig::Curve-GTOT}
\end{figure*}

\section{Performance Evaluation}
\label{sec::evaluation}
To validate the effectiveness of our quality-aware Feature Aggregation Network (FANet), we test it on two large-scale RGB-T benchmarks, including GTOT dataset~\cite{Li16tip} and RGBT234
dataset~\cite{li2019rgb} and analyze the tracking performance.

\subsection{Evaluation Setting}

\begin{figure*}[t]
  \centering
  \includegraphics[width=\textwidth]{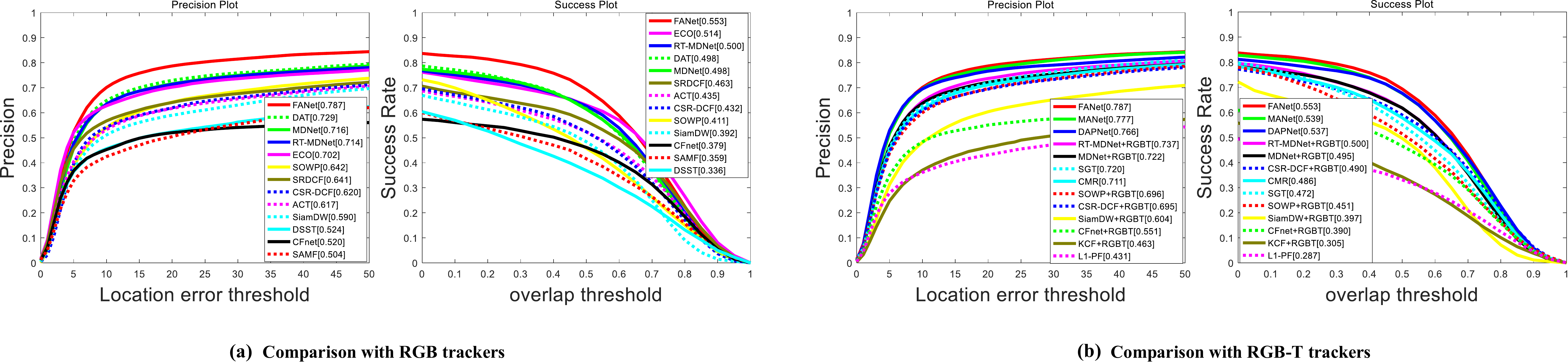}\\
  \caption{The evaluation results on RGBT234 dataset. The representative scores of PR/SR is presented in the legend. We separate RGB and RGB-T trackers in (a) and (b). }\label{fig::Curve-RGBT234}
\end{figure*}

\begin{table*}[t]\footnotesize 
\renewcommand\arraystretch{1.5}
\setlength{\belowcaptionskip}{0.2cm}
\caption{Attribute-based PR/SR scores (\%) on RGBT234 dataset against with eight RGB-T trackers. The best and second results are in \textcolor{red}{red} and \textcolor{green}{green} colors, respectively.}
\centering
\begin{tabular}{ c | c  c  c  c  c  c  c  c | c }
	\hline
   & SOWP+RGBT &CFNet+RGBT &CMR &L1-PF &CSR-DCF+RGBT & DAPNet & SGT   &RT-MDNet+RGBT & FANet \\\hline
 NO &{86.8}/53.7 & 76.4/56.3 & \textcolor{green}{89.5}/61.6 & 56.5/37.9 & 82.6/{60.0} & \textcolor{red}{90.0}/\textcolor{green}{64.4} & {87.7}/55.5 & 86.7/{61.1} & 88.2/\textcolor{red}{65.7}  \\

 PO & 74.7/48.4 & 59.7/41.7 & 77.7/53.5 & 47.5/31.4 & 73.7/{52.2} & \textcolor{green}{82.1}/\textcolor{green}{57.5} & {77.9}/51.3 & 81.3/55.4 & \textcolor{red}{86.6}/\textcolor{red}{60.3} \\

 HO & 57.0/37.9 & 41.7/29.0 & 56.3/37.7 & 33.2/22.2 &59.3/40.9 & \textcolor{green}{66.0}/\textcolor{green}{45.7} & 59.2/39.4 & {60.3}/{39.7} & \textcolor{red}{66.5}/\textcolor{red}{45.8} \\

 LI & {72.3}/46.8 & 52.3/36.9 &74.2/49.8 & 40.1/26.0 & 69.1/{47.4} &\textcolor{green}{77.5}/\textcolor{green}{53.0} & 70.5/46.2 & 71.3/47.4 & \textcolor{red}{80.3}/\textcolor{red}{54.8} \\

 LR & 72.5/46.2 & 55.1/36.5 & 68.7/42.0 & 46.9/27.4 & 72.0/47.6 &75.0/\textcolor{green}{51.0} &\textcolor{green}{75.1}/47.6 &{74.6}/{48.8} & \textcolor{red}{79.5}/\textcolor{red}{53.2} \\

 TC & 70.1/44.2 & 45.7/32.7 & 67.5/44.1 &37.5/23.8 & 66.8/46.2 & \textcolor{red}{76.8}/\textcolor{green}{54.5} & {76.0}/47.0 & 72.1/{50.8} & \textcolor{green}{76.6}/\textcolor{red}{55.1}  \\

 DEF & 65.0/46.0 & 52.3/36.7 & 66.7/47.2 & 36.4/24.4 & 63.0/46.2 & \textcolor{green}{71.7}/\textcolor{green}{51.9} & {68.5}/{47.4} & 64.8/46.4 & \textcolor{red}{72.2}/\textcolor{red}{52.7} \\

 FM & {63.7}/38.7 & 37.6/25.0 & 61.3/38.2 & 32.0/19.6 & 52.9/35.8 & {67.0}/\textcolor{red}{44.5} & \textcolor{green}{67.7}/{40.2} & 64.6/40.2 &\textcolor{red}{68.1}/\textcolor{green}{43.8} \\

SV &66.4/40.4 & 59.8/43.3 & 71.0/49.3 & 45.5/30.6 & 70.7/49.9 & \textcolor{green}{78.0}/\textcolor{green}{54.2} & 69.2/43.4 &{73.9}/{50.5} & \textcolor{red}{78.5}/\textcolor{red}{56.3} \\

 MB & 63.9/42.1 & 35.7/27.1 & 60.0/42.7 & 28.6/20.6 & 58.0/42.5 & \textcolor{green}{65.3}/\textcolor{green}{46.7} & 64.7/43.6 & {63.5}/{45.1} &\textcolor{red}{70.0}/\textcolor{red}{50.3} \\

 CM & 65.2/43.0 & 41.7/31.8 & 62.9/44.7 & 31.6/22.5 & 61.1/44.5 & \textcolor{green}{66.8}/\textcolor{green}{47.4} & {66.7}/45.2 & 64.6/{44.7} & \textcolor{red}{72.4}/\textcolor{red}{52.3} \\

 BC & 64.7/41.9 & 46.3/30.8 & 63.1/39.7 & 34.2/22.0 & 61.8/41.0 & \textcolor{green}{71.7}/\textcolor{green}{48.5}& {65.8}/41.8 & {64.2}/{40.8} &\textcolor{red}{75.7}/\textcolor{red}{50.3}\\\hline

 ALL &69.6/45.1 & 55.1/39.0 & 71.1/48.6 & 43.l1/28.7 & 69.5/49.0 & \textcolor{green}{76.6}/\textcolor{green}{53.7} & 72.0/47.2 & {73.7}/{50.0} & \textcolor{red}{78.7}/\textcolor{red}{55.3} \\\hline
\end{tabular}
\label{tb::AttributeResults}
\end{table*}

{\flushleft \bf Datasets}.
We evaluate our FANet on two large-scale benchmark datasets, i.e., GTOT dataset~\cite{Li16tip} and RGBT234 dataset~\cite{li2019rgb}.

GTOT is a standard benchmark dataset for RGBT tracking.
It contains 50 RGBT video sequences under different scenes and conditions, and the target box for each frame is annotated.
In addition, it is annotated with 7 attributes for analyzing the attributed-sensitive performance of RGBT tracking approaches. 
The RGBT234 dataset is currently the largest RGBT tracking dataset, which is extended from the rgbt210 dataset ~\cite{Li17rgbt210}. 
It contains 234 rgbt videos with a frame count of approximately 234,000 and a maximum video pair frame size of 8000. 
In order to analyze the attribute-based performance of different tracking algorithms, 12 attributes are used for annotation.

\begin{figure*}[t]
  \centering
  \includegraphics[width=1\textwidth]{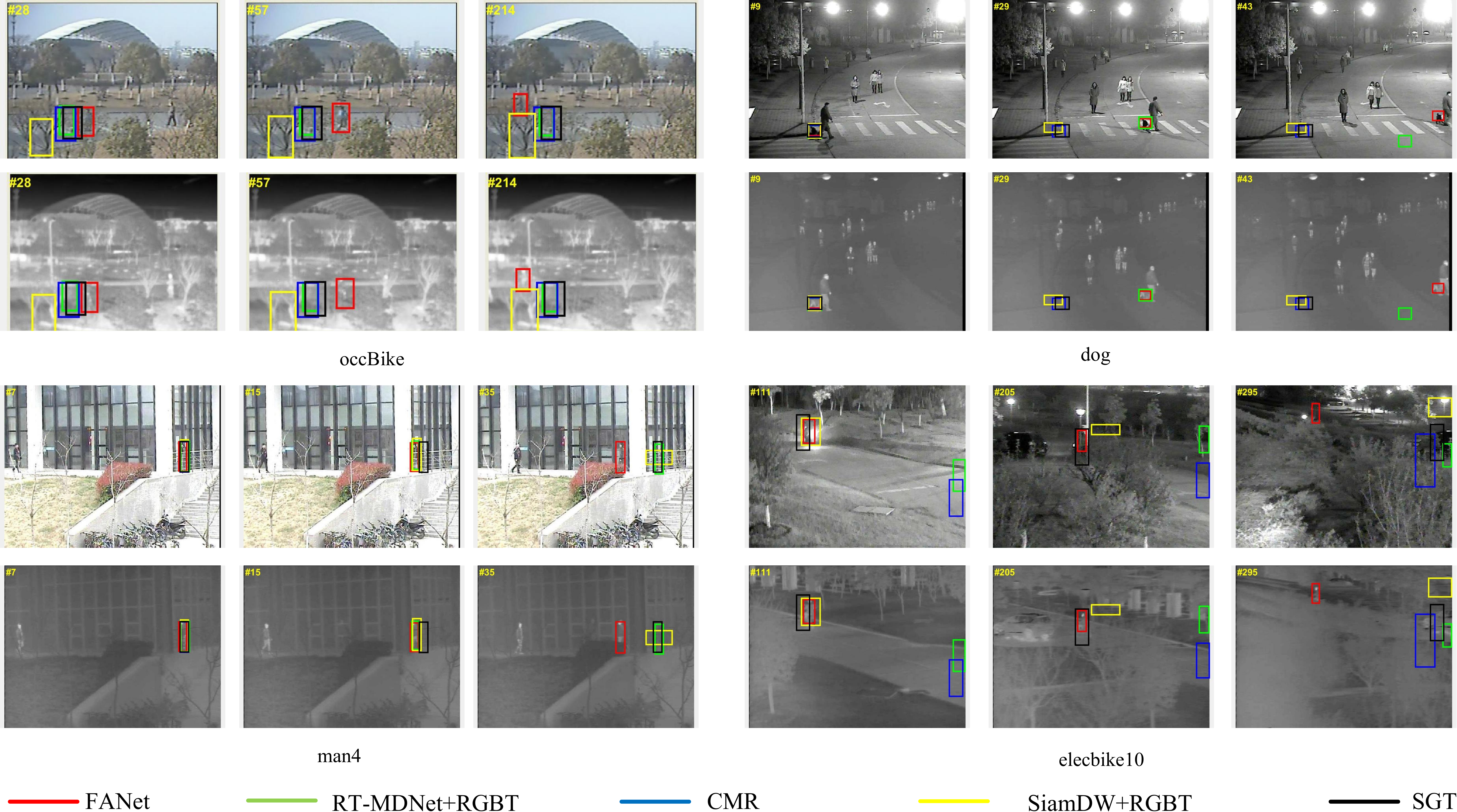} \\
  \caption{Qualitative comparison of our FANet versus four state-of-the-art RGB-T trackers on four video sequences. }\label{fig::VisualResults}
\end{figure*}

{\flushleft \bf Evaluation metrics}.
On these two datasets, we use success rate (SR) and precision rate (PR) for quantitative performance evaluation.
PR is the percentage of frames whose output location is within a threshold distance of the ground truth. 
SR is the percentage of frames whose overlap rate between the output boundary box and the ground truth boundary box is largar than a threshold.
 The threshold of PR is set to 5 and 20 pixel for GTOT and RGBT234 datasets respectively (because the target object in the GTOT dataset is generally small).
And the area under the curves of success rate is adopted to represent SR for quantitative performance evaluation.

{\flushleft \bf Baseline}.
In order to verify the validity of our RGBT tracking algorithm, we compare our method with 29 popular trackers, some of which are from GTOT and RGBT234 benchmark.

\subsection{Evaluation on GTOT Dataset}

{\flushleft \bf Comparison with RGB trackers}.
To verify the superiority of the proposed RGBT tracking method compared to RGB trackers, we first evaluate our method with 12 state-of-the-art RGB trackers, including MDNet~\cite{MDNet15cvpr}, DAT~\cite{Pu2018Deep}, ECO~\cite{ECO17cvpr}, RT-MDNet~\cite{RT-MDNet18eccv}, SRDCF~\cite{danelljan2015learning}, BACF~\cite{Galoogahi2017Learning}, ACFN~\cite{Choi2017Attentional}, ACT~\cite{chen2018real}, SiamDW~\cite{Zhipeng2019Deeper}, CFnet~\cite{CFNet17cvpr} and KCF~\cite{henriques2015high}, SiameseFC~\cite{Bertinetto2016Fully}. 
Fig.~\ref{fig::Curve-GTOT}(a) demonstrates the effectiveness of introducing thermal information in visual tracking. 
In particular, our tracker outperforms DAT, RT-MDNet and ECO with 12.0\%/11\%, 14.6\%/11.5\% and 12.1\%/9.7\% in PR/SR, respectively.

{\flushleft \bf Comparison with RGBT trackers}.
On the GTOT dataset~\cite{Li16tip}, we compare our approach with 16 state-of-the-art trackers, some of which are from GTOT benchmark. 

Since there are few RGBT trackers~\cite{yang2019icip,zhu2019dense,Li16mmm,Liu12infosci,Li16tip,Li17rgbt210,Li18eccv}, some RGBT tracking methods are derived by extending the RGB method. Specifically, we implement rgbt tracking by directly concatenate RGB and thermal features or considering thremal as an additional channel to input RGB tracking. 
In particular, RT-MDNet+RGBT, DAT+RGBT, SiamDW+RGBT and MDNet+RGBT are extended RGBT tracking algorithms that directly concatenate RGB and thermal channels as input.  
The top 12 are DAPNet~\cite{zhu2019dense}, DAT~\cite{Pu2018Deep}+RGBT, CMR~\cite{Li18eccv}, MDNet~\cite{MDNet15cvpr}+RGBT, SGT~\cite{Li17rgbt210}, LTDA~\cite{yang2019icip}, Struck~\cite{Stuck11iccv}+RGBT, RT-MDNet~\cite{RT-MDNet18eccv}+RGBT, CFnet~\cite{CFNet17cvpr}+RGBT and SiamDW~\cite{Zhipeng2019Deeper}+RGBT,JSR~\cite{Liu12infosci} and L1-PF~\cite{Wu11icif}).
Fig.~\ref{fig::Curve-GTOT}(b) shows that our tracker is superior to other state-of-the-art trackers on the GTOT dataset. 
Specifically, our method improves
the CMR, RT-MDNet+RGBT and SGT by a large margin, \emph{i.e.}, 6.4\%/8.5\%, 5.2\%/5.9\% and 4.0\%/10.0\% in PR/SR. 
In addition, our method is slightly better than DAPNet~\cite{zhu2019dense} 0.9\%/2.1\% in PR/SR, and our algorithm runs 6 times faster. 
The overall promising performance of our method can be explained by the fact that FANet makes fully use of hierarchical deep features and RGBT information to well handle the challenges of significant appearance changes and adverse environmental conditions.


\subsection{Evaluation on RGBT234 Dataset}

{\flushleft {\bf Overall performance}}.
For more comprehensive evaluation, we report the evaluation results on the RGBT234 dataset~\cite{li2019rgb}, as shown in Fig.~\ref{fig::Curve-RGBT234}. 
The comparison trackers include 12 RGB ones, i.e., (DAT~\cite{Pu2018Deep}, RT-MDNet~\cite{RT-MDNet18eccv}, ACT~\cite{chen2018real} SiamDW~\cite{Zhipeng2019Deeper}, ECO~\cite{ECO17cvpr}, MDNet~\cite{MDNet15cvpr}, SOWP~\cite{Kim15iccv}, SRDCF~\cite{danelljan2015learning}, CSR-DCF~\cite{dcf-csr16cvpr}, DSST~\cite{danelljan2014accurate}, CFnet~\cite{CFNet17cvpr} and SAMF~\cite{li2014scale}),
and 12 RGBT ones, i.e., (MANet~\cite{li2019multi}, MDNet+RGBT, CMR~\cite{Li18eccv},SGT~\cite{Li17rgbt210}, SOWP+RGBT, CSR-DCF+RGBT, MEEM~\cite{MEEM14eccv}+RGBT, CFnet+RGBT, KCF~\cite{henriques2015high}+RGBT and L1-PF~\cite{Wu11icif}). 
From the results, we can see that the performance of our FANet clearly better than the advanced RGB and RGBT methods in all metrics. 
It demonstrate the importance of thermal information and feature aggregations proposed in our method. 
In particular, our FANet achieves 5.8\%/5.5\% performance gains in PR/SR over the second best RGB tracker DAT, achieves 1.0\%/1.4\% gains over the second best RGBT tracker MANet, and our algorithm runs 15 times faster than MANet. 

{\flushleft {\bf Attribute-based performance}}.
We also report the attribute-based results of our FANet versus other state-of-the-art RGB-T trackers, including L1-PF, SOWP+RGBT, CSR-DCF+RGBT, CFNet+RGBT, RT-MDNet+RGBT, SGT, CMR and DAPNet, as shown in Table~\ref{tb::AttributeResults}. 
The attributes include no occlusion (NO), partial occlusion (PO), heavy occlusion (HO), low illumination (LI), low resolution (LR), thermal crossover (TC), deformation (DEF), fast motion (FM), scale variation (SV), motion blur (MB), camera moving (CM) and background clutter (BC). 
The results show that our method performs the best in terms of most challenges except for NO, TC and FM. 
It demonstrates the effectiveness of our FANet in handling the sequences with the appearance changes and adverse conditions. 
The following major observations and conclusions can be drawn from Table~\ref{tb::AttributeResults}.

First, although CMR and DAPNet perform well when no occlusion occurs in PR, their performance drop a lot when partial or heavy occlusions happen. 
Our FANet keeps high tracking performance with partial or heavy occlusions, which suggests that the aggregated hierarchical features can improve the tracking robustness in presence of occlusions. 
Second, in adverse lighting conditions and low resolution, FANet outperforms all other trackers. 
It can be explained that adaptive incorporation from RGB and thermal information can boost tracking performance significantly (\emph{e.g.} FANet versus RT-MDNet+RGBT). 
And our strategy to predict the modality weights is more robust than other existing methods (\emph{e.g.}, FANet versus SGT). 
Third, our framework is robust to the significant appearance changes of target object and the effects of cluttered background by observing its performance under deformation, scale variation, camera moving and background clutter. 
Finally, unsatisfying results are usually generated in fast motion video sequences by our FANet. 
It might be that the rapid movement of the target object is beyond the scope of the searching windows.

{\flushleft {\bf Qualitative performance}}.
The qualitative comparison of our FANet versus 4 state-of-the-art RGB-T trackers on 4 video sequences is presented in Fig.~\ref{fig::VisualResults}, including RT-MDNet~\cite{RT-MDNet18eccv}+RGBT, SGT~\cite{Li17rgbt210}, CMR~\cite{Li18eccv} and SiamDW~\cite{Zhipeng2019Deeper}+RGBT. 
Overall, our method is effective in handling the challenges of occlusion, background clutter, appearance variation, low resolution and low illumination variation. 
For example, in Fig.~\ref{fig::VisualResults} (occBike), our method performs well in presence of partial and heavy occlusions and background clutter while other trackers lose the target when occlusion happens. 
In Fig.~\ref{fig::VisualResults} (dog), the target object moves so fast that most algorithms can't track it, and only our method can do it. 
In Fig.~\ref{fig::VisualResults} (man4), we can see that our method has a relatively good treatment for heavy occlusion. 
In Fig.~\ref{fig::VisualResults} (elecbike10), the \emph{elecbike10} is partly invisible in visible source but the thermal images can provide relatively reliable information to distinguish the target object from the background. 
Moreover, in the case of low resolution and occlusions, all compared methods lose the targets, while our FANet method is able to track it robustly by adaptively incorporating useful information from these two modalities. 
From the above qualitative comparison, we can find that our algorithm is more capable of dealing with complex challenges in real scenes.

\begin{table}[t]\footnotesize
\renewcommand\arraystretch{1.5}
\caption{PR/SR scores (\%) of different variants induced from our network on GTOT and RGBT234 datasets. }
\centering
\setlength{\tabcolsep}{1.5mm}{
\begin{tabular}{c|c| c c c c   }
\hline
 && RT-MDNet+RGBT & FANet-FA & FANet-MA & FANet  \\
   \hline
{\bf GTOT}    &	PR &83.9 & 88.5 & 88.7 & \bf89.1   \\
              &   SR &66.9& 72.7 &72.6 & \bf72.8  \\
\hline
{\bf RGBT234} & PR & 73.7 & 77.0 &  76.7 & \bf78.7\\
              & SR & 50.0 & 53.0 &  54.0 & \bf55.3\\
\hline
\end{tabular}}
\label{tb::component-analysis}
\end{table}

\begin{figure}[t]
  \centering
  \includegraphics[width=\columnwidth]{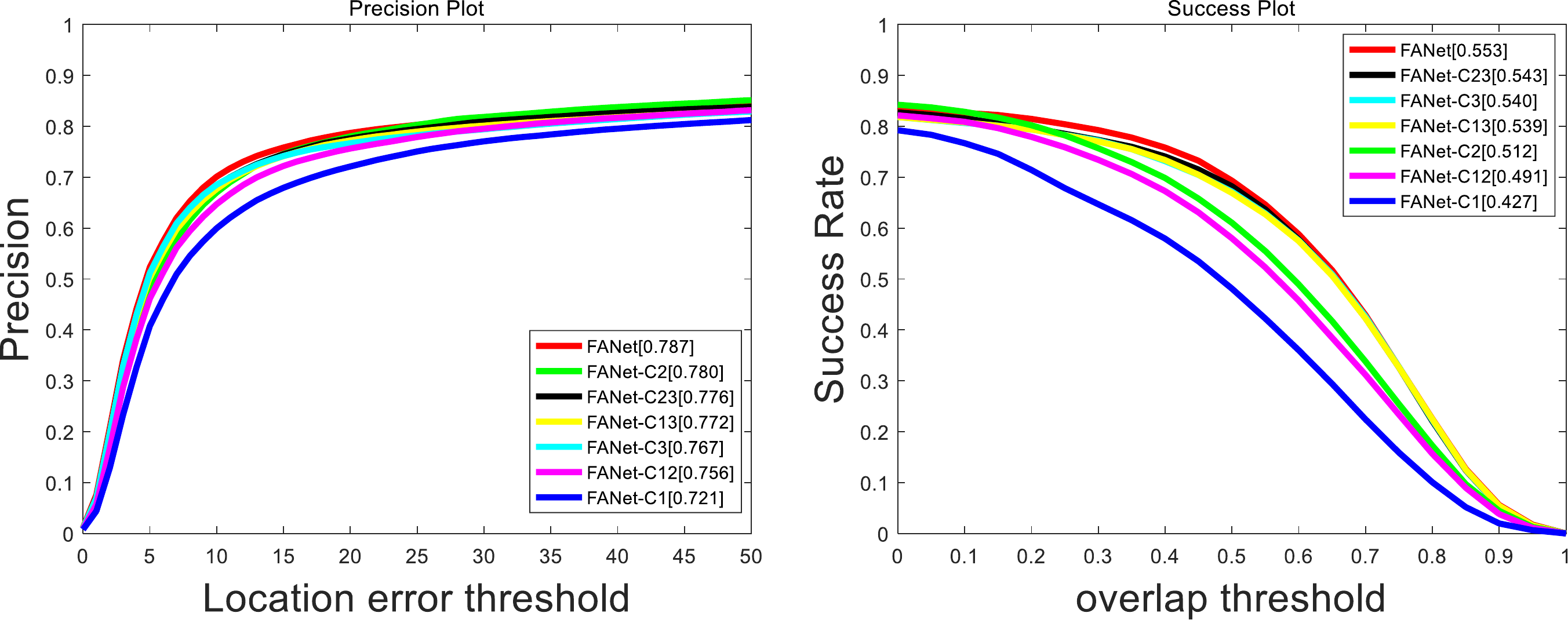} \\
  \caption{Performance evaluation using different convolutional layers on RGBT234 dataset. }\label{fig::Results_features}
\end{figure}

\begin{table*}[t]\footnotesize
\renewcommand\arraystretch{1.5}
\caption{PR(\%) and SR(\%) scores of our algorithm on GTOT and RGBT234 comparing with different fusion stages.}
\centering
\setlength{\tabcolsep}{2mm}{
\begin{tabular}{c| c |  c  c   c c |  c }
	\hline
   & & RT-MDNet& RT+RGBT-EARLY& RT+RGBT-MID& RT+RGBT-LATE& FANet\\\hline
GTOT& PR& 74.5& 84.6& 79.4&  82.9& \bf89.1  \\
    & SR& 61.3& 68.3& 67.2&  66.7& \bf72.8 \\\hline

RGBT234& PR& 71.4& 73.7& 72.2&  73.4& \bf78.7 \\
       & SR& 50.0& 50.0& 49.3&  48.3& \bf55.3 \\\hline
\end{tabular}}
\label{tb:baseline}
\end{table*}
\subsection{Analysis of Our Network}
{\flushleft \bf Ablation study}.
In order to prove the importance of the main components, we apply two special versions of our method for comparative analysis and evaluation them on both GTOT and RGBT234 datasets. 
These two variants are: 1) FANet-FA, that only uses the layer weights and removes the adaptive aggregation subnetwork; 
and 2) FANet-MA, that only uses the modality weights to integrate the feature of Conv3 of two modalities and removes the hierarchical feature aggregation module. 
Table~\ref{tb::component-analysis} presents the comparison results. 
The performance of our FANet-FA, FANet-MA are obviously better than baseline(RT-MDNet+RGBT), which proves the validity of our sub-networks. 
In addition, the performance of our FANet over FANet-FA and FANet-MA demonstrates the effectiveness of the proposed entire network.

{\flushleft \bf Aggregation Analysis}. 
To further verify the validity of our aggregation method, we extend RT-MDNet algorithm for RGBT tracking of dual-modal inputs, and have three aggregate forms. 
The first one is that we directly concatenate two modalities of data channels to form 6 channels of input data, then input the original RT-MDNet algorithm for tracking, which is named RT+RGBT-EARLY (equivalent to RT-MDNet+RGBT). 
The second one is that we concatenate the feature maps of the modalities at Conv1, and named RT+RGBT-MID. 
The last one is that we extract the convolution features of the two modalities separately, and concatenate the feature maps of Conv3, named RT+RGBT-LATE. 
Table~\ref{tb:baseline} shows the results compared with the baseline on GTOT and RGBT234 datasets. 
From the Table ~\ref{tb:baseline}, we can see that our method that aggregates two modalities is obviously superior than RT-MDNet, which proves that the fusion of two modal information is beneficial to visual tracking. 
In addition, on GTOT and RGBT234 datasets, the FANet outperforms the best baseline method with 4.5\%/4.5\%, 5.0\%/5.3\% in PR/SR, respectively. 
It shows that our FANet is very effective for representing and fusing different modalities in visual tracking.

{\flushleft \bf Feature analysis}.
To analyze the effectiveness of hierarchical feature aggregation in the proposed network, we compare the tracking results using deep features from different convolutional layers on the RGBT234 dataset. 
We first use each individual convolutional layer (C1, C2 and C3) to represent objects in the network. 
The tracking performance is better when a deeper layer is used, which can be attributed to the discriminative semantic abstraction from deeper layers.  
Moreover, we also
use combinations of C1, C2 and C3 (C12, C13 and C23) to represent objects in the proposed network. 
Fig.~\ref{fig::Results_features} shows the evaluation results on the RGBT234 dataset.
The superior performance of our FANet over others demonstrates the effectiveness of the used hierarchical features. 
From PR/SR scores, we can find that FANet-C13 and FANet-C23 methods perform better than FANet-C1, FANet-C2 and FANet-C3. 
The results show that using multi-layer network features simultaneously can improve tracking performance.

\begin{table*}[t]\footnotesize
\renewcommand\arraystretch{1.5}
\caption{Performance and runtime of our FANet against the baseline methods on RGBT234 datasets.}
\centering
\setlength{\tabcolsep}{2mm}{
\begin{tabular}{c| c  |   c c c  c c |  c }
	\hline
   &   & MDNet+RGBT &MANet &DAPNet &CMR &SGT  & FANet \\
    \hline

RGBT234&PR  &72.2 &77.7 &76.6 &71.1 &72.0 & \bf78.7 \\
 &SR       &49.5 &53.9 &53.7 &48.6 &47.2 &\bf55.3\\\hline
 &FPS        & 3   &1.5  &2    &8    &5    &\bf19\\\hline
\end{tabular}}
\label{tb::runtime}
\end{table*}

\begin{figure*}[t]
  \centering
   \includegraphics[width=1\textwidth]{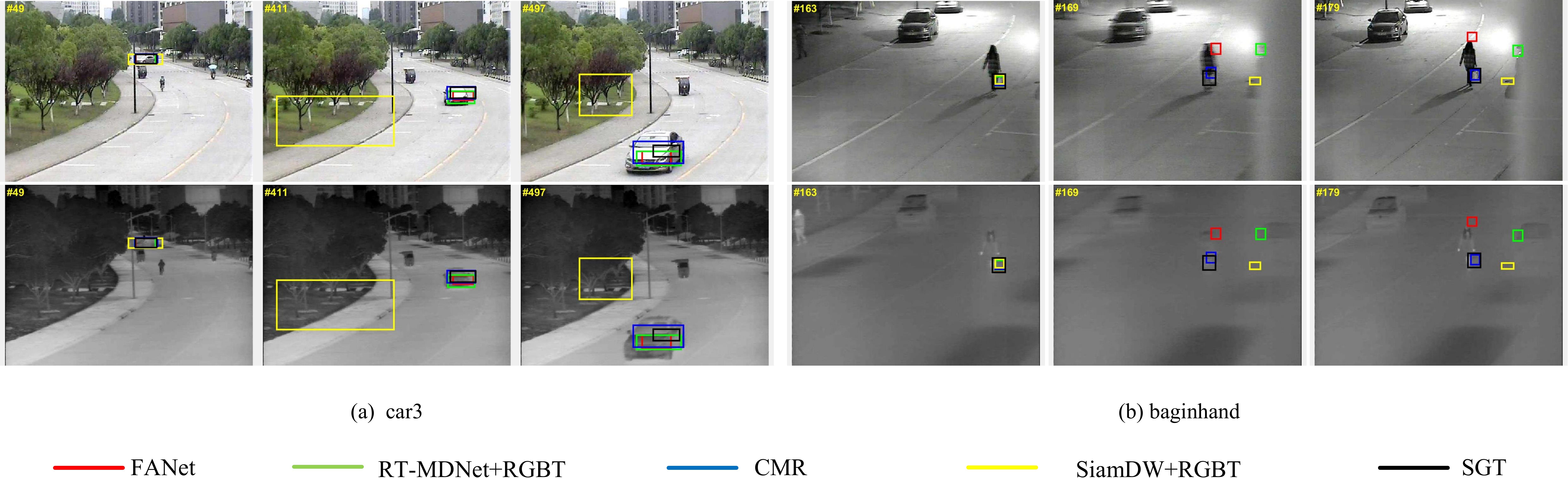} \\
  \caption{Failure cases on the car3 and  baginhand (RGBT234). }\label{fig::failure-cases}
\end{figure*}

{\flushleft \bf Runtime analysis}.
Finally, we present the runtime of our
FANet against the state-of-the-art trackers, MDNet~\cite{MDNet15cvpr}+RGBT, MANet~\cite{li2019multi}, DAPNet~\cite{zhu2019dense}, CMR~\cite{Li18eccv}, SGT~\cite{Li17rgbt210} with their tracking performance on the RGBT234 dataset in Table~\ref{tb::runtime}. 
Our implementation is on the platform of PyTorch0.41 with 2.1 GHz Intel(R) Xeon(R) CPU E5-2620 and NVIDIA GeForce GTX 2080Ti GPU, and the average tracking speed is 19 FPS.
Overall, the results demonstrate that our framework clearly outperforms
other tracking methods. 

{\flushleft \bf Failure cases}
We show some cases of tracking failure by FANet in FIg.~\ref{fig::failure-cases}. 
The $car3$ sequence in Fig.~\ref{fig::failure-cases}(a), when the object scale variation too much, the proposed method fails to located the target. 
The reason for the failure might be that our tracker only learns a linear scale regression estimation parameter in the first frame, which is difficult to cope with large-scale variations. 
The $baginhand$ sequence in Fig.~\ref{fig::failure-cases}(b), the sudden camera shake can cause our tracker to not successfully track the target object. 
The sudden large shake of cameras can cause our tracker to not successfully track the target object. 
There are two possible reasons.
1) Sudden camera shake will make the target exceed our limited search range.
2) Violent camera shake will blur the image, and blurred appearance and sudden changes would make our classifier difficult to distinguish the target object from the background.

\section{Conclusion}
\label{sec::conclusion}
In this paper, we propose an end-to-end trained FANet for robust RGB-T tracking. 
FANet aggregates hierarchical multi-resolution feature maps to handle significant appearance changes caused by deformation, low illumination, background clutter and occlusion within each modality, and predict the reliability degrees to address the problem of uncertainties in different modalities. 
Extensive experiments on large-scale benchmark datasets show that the FANet gains excellent tracking result.
In future work, we will develop more efficient and effective scale handling methods, such as IOUNet~\cite{jiang2018acquisition}, into our framework for powerful RGBT tracking.



\ifCLASSOPTIONcaptionsoff
  \newpage
\fi



%

\bibliographystyle{ieeetran}
\bibliography{FANet}

%

\end{document}